\documentclass[letterpaper, 10 pt, conference]{ieeeconf}
\IEEEoverridecommandlockouts
\usepackage[vlined, ruled, boxed, linesnumbered]{algorithm2e}

\SetCommentSty{mycommfont}

\SetKwProg{Function}{}{}{}

\usepackage{graphicx}
\usepackage{graphics}
\usepackage{times}
\usepackage{amsmath}
\usepackage{amssymb}
\usepackage{float}
\usepackage{url}
\usepackage{subfigure}
\usepackage{caption}
\usepackage{multirow}
\usepackage{gensymb}
\usepackage{epstopdf}
\usepackage{wrapfig}
\usepackage{setspace} 
\usepackage[noend]{algpseudocode}
\usepackage{color}
\usepackage{cite}
\usepackage{booktabs}
\usepackage{stackengine}
\usepackage[font={small}]{caption}

\usepackage{tikz}
\usepackage{textcomp}
\usepackage{lipsum}

\title{\LARGE \bf
Graph2Nav: 3D Object-Relation Graph Generation to Robot Navigation
}

\author{Tixiao Shan$^{*}$, Abhinav Rajvanshi$^{*}$, Niluthpol Mithun, and Han-Pang Chiu
\thanks{$^{*}$ The first two authors have equal contribution. All authors are with Center for Vision Technologies, SRI International, Princeton, NJ 08550, USA. The contact author is Han-Pang Chiu ({\tt\small han-pang.chiu@sri.com}).}%
}

\begin{document}

\maketitle
\thispagestyle{empty}
\pagestyle{empty}


\begin{abstract}
We propose Graph2Nav, a real-time 3D object-relation graph generation framework, for autonomous navigation in the real world. Our framework fully generates and exploits both 3D objects and a rich set of semantic relationships among objects in a 3D layered scene graph, which is applicable to both indoor and outdoor scenes. It learns to generate 3D semantic relations among objects, by leveraging and advancing state-of-the-art 2D panoptic scene graph works into the 3D world via 3D semantic mapping techniques. This approach avoids previous training data constraints in learning 3D scene graphs directly from 3D data. We conduct experiments to validate the accuracy in locating 3D objects and labeling object-relations in our 3D scene graphs. We also evaluate the impact of Graph2Nav via integration with SayNav, a state-of-the-art planner based on large language models, on an unmanned ground robot to object search tasks in real environments. Our results demonstrate that modeling object relations in our scene graphs improves search efficiency in these navigation tasks.

\end{abstract}


\section{Introduction}
\label{sec::introduction}


3D scene graphs \cite{3DSG19, Kimera, Hydra, 3DSG, SGraph, OpenGraph, 3DSGOntologies, HSG2Robotics, Wald20, Wu21, Wu23}, where nodes depict the objects and edges characterize the relations among objects, have become popular high-level representations of 3D large-scale environments. The main advantage of a 3D scene graph over other object-based 3D scene representations is its capability also to represent semantic relationships (e.g. ``beside", ``in front of", ``on top of") among objects. These relationships are useful to many downstream applications, such as scene manipulation \cite{imagemanipulation, GraphTo3D} and task planning \cite{taskography}. 

Leveraging 3D scene graphs to robot navigation has also emerged as a promising research field with impressive performance \cite {SayPlan,SayNav,GraphPolicy}. For example, \cite {SayPlan} and \cite{SayNav} utilize knowledge from Large Language Models (LLMs) for generating high-level plans to navigation. 3D scene graphs are used as inputs to LLMs, for ensuring the generated plans adhere to the 3D objects with their locations and properties presented by the perceived environment in which the robot operates. 

However, there are two major limitations in current 3D scene graph generation methods which hinder the growth of this field. First, semantic relationships among objects are rarely explored in works to real-time 3D scene graph generation \cite{Kimera, Hydra, SayNav}, limiting their capabilities to robot navigation tasks. Second, almost all 3D scene graph generation datasets \cite{Wu21, Wu23} are based on indoor scenes. Therefore, these works are not applicable to outdoor environments.     
\begin{figure}[ht]
	\centering
	\includegraphics[width=.48\textwidth]{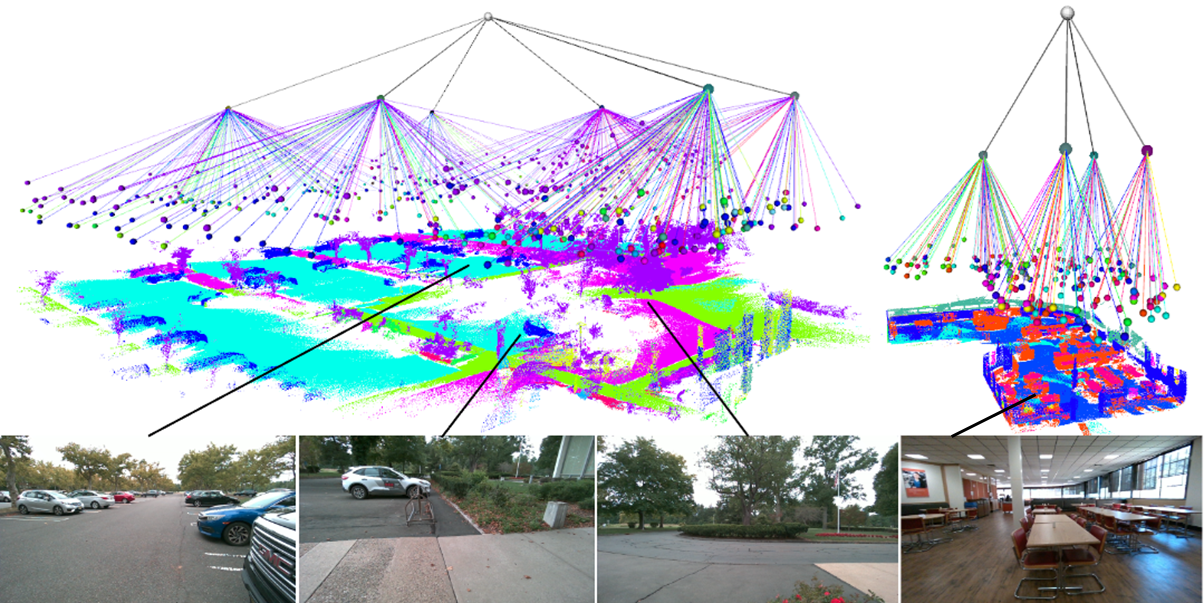}
	\caption{3D scene graphs constructed using Graph2Nav for outdoor (left) and indoor (right) scenes. The graph includes a hierarchy (from top to bottom): a test site, regions, and objects. The figure also shows examples of 2D input images and the 3D point clouds generated by Graph2Nav. Note we omit text labels of objects and object-relations in 3D scene graphs for better visualization.}
	\label{fig::Graph2Nav}
	\vspace{-5mm}
\end{figure}

In this paper, we present Graph2Nav (Figure \ref{fig::Graph2Nav}), a novel real-time 3D object-relation graph generation framework that addresses these limitations to robot navigation. Our framework fully generates and exploits both 3D objects and a rich set of semantic relationships among objects in a 3D scene graph, which is applicable to both indoor and outdoor scenes. Instead of direct 3D scene graph generation, our framework generates a 2D object-relation graph from each key video frame via a graph prediction network. It then utilizes 3D simultaneous localization and mapping (SLAM) techniques to continuously merge newly generated 2D graphs into a global 3D scene graph. This approach avoids training data constraints in prior works in learning 3D scene graphs, by leveraging existing large-scale indoor/outdoor 2D training datasets (containing object-relation labels) \cite{SGSurvey} for graph prediction.      

Our main contributions are summarized as follows:
\begin{itemize}
	\item We propose Graph2Nav, a new real-time 3D object-relation graph generation framework that combines strengths of 2D object-relation graphs and 3D semantic mapping techniques. 
	\item We integrate Graph2Nav with a LLM-based planner (SayNav\cite{SayNav}) and enhance its graph generation capabilities across both indoor and outdoor scenes. 
	\item We evaluate the impact from our 3D scene graphs via SayNav on a UGV (unmanned ground vehicle) to object search tasks in real environments. To the best of our knowledge, this is the first autonomous robot that uses 3D scene graphs to ground LLMs for navigation in real unknown environments.   
\end{itemize}

\begin{figure*}[ht]
	\centering
	\includegraphics[width=.99\textwidth]{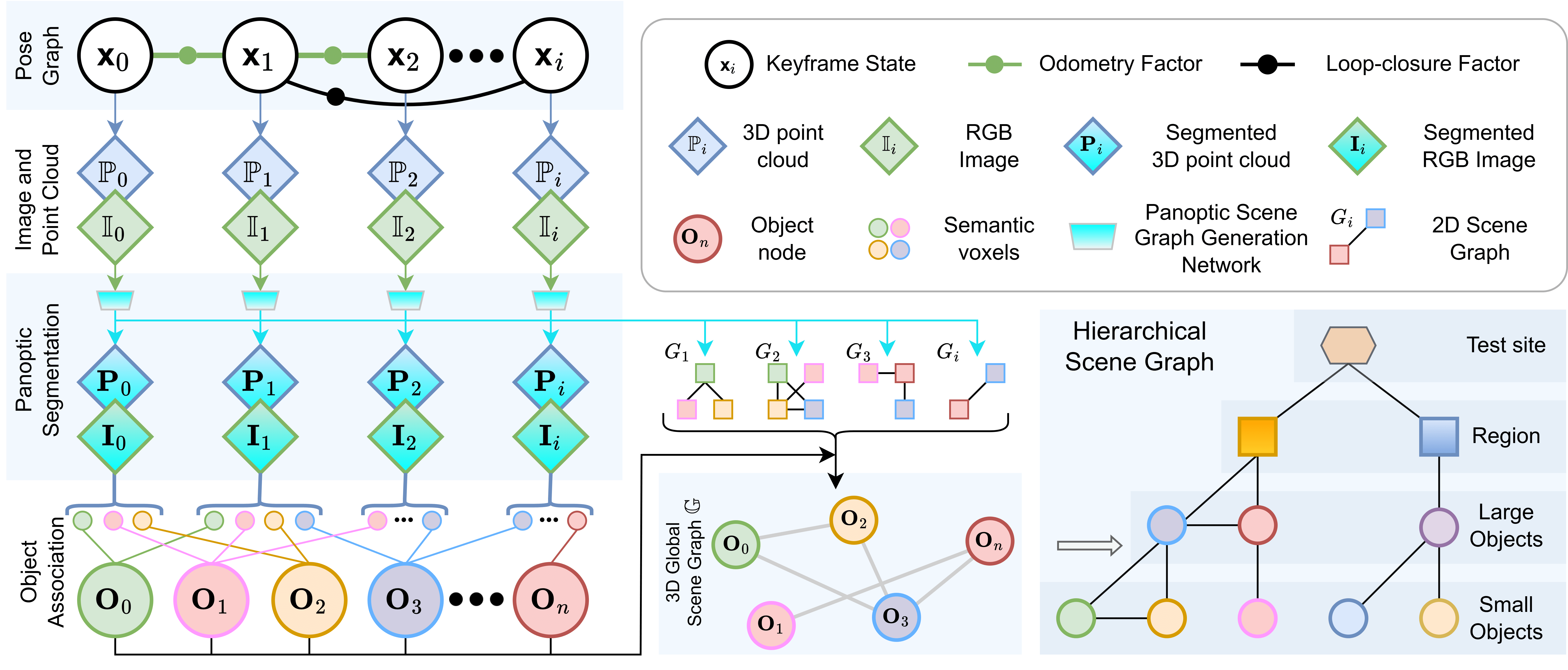}
	\caption{The process flow diagram for Graph2Nav. A pose graph-based SLAM system is utilized to provide real-time pose estimations for received image and point cloud data. Semantic objects and relations are then extracted from them via a panoptic scene graph generation network. At last, a consistent global 3D scene graph is generated by continuously merging newly observed objects and relations.}
	\label{fig::system-arch}
	\vspace{-5mm}
\end{figure*}

\section{Related Work}
\label{sec::relatedwork}
\subsection{3D Scene Graphs}
\label{sec::3DSG}
There are two major groups of works in 3D scene graph generation. The first group \cite{3DSG19, Kimera, Hydra, 3DSG, SGraph, OpenGraph, 3DSGOntologies, HSG2Robotics} typically combines traditional SLAM methods and novel geometric deep learning techniques, such as Graph Neural Networks (GNN) and Transformers. The resulting 3D scene graph is a layered graph which represents spatial concepts (nodes) at multiple levels (such as objects, places,
rooms, and buildings) with their relations (edges). The layered structure is defined either manually \cite{3DSG19, Kimera, Hydra, 3DSG, SGraph, OpenGraph} or via learning techniques \cite{3DSGOntologies, HSG2Robotics}. Due to the lack of outdoor datasets, most of these works are only applicable to indoor scenes. In addition, all these works only label edges across layers (such as objects ``inside" a room) in the graph. None of these works explore the semantic relations (such as ``beside" and ``in front of") as edges among objects. 

The second group \cite{Wald20, Wu21, Wu23} focuses on learning a 3D flat (one-layer) scene graph directly from 3D point clouds via GNN and instance segmentation techniques. The 3D point cloud is built either from RGBD sequences \cite{Wu21} or RGB sequences \cite{Wu23}. These works explicitly learn object relations without the layered structure in the 3D scene graph. However, they are limited to small-scale indoor scenes (such as a room) due to training data constraints.

Our work bridges the gap between these two groups in 3D scene graph generation. Specifically, we focus on learning 3D semantic relations among objects in the 3D layered graph structure, which is applicable to both indoors and outdoors.  

\subsection{2D Object-Relation Graphs}
\label{sec::2DSG}
In computer vision, there is a rich history in learning a 2D object-relation graph (which is also called a scene graph) from a single image. There are many existing large-scale benchmark datasets with labeled 2D object-relation graphs for both indoor and outdoor images. A comprehensive survey in this field can be found in \cite{SGSurvey}. The typical approach is to detect objects followed by the prediction of their pairwise relationships on an image via a GNN or a transformer. However, using only detected objects cannot cover the full scene of an image (such as the room layout). Therefore, these detection-based methods cannot fully fulfill the requirements to our application. 

Recently, there are works \cite{PanopticSG, PanopticSGL} in utilizing panoptic segmentation \cite{PanopticSegmentation} to replace the object detection process in 2D object-relation graph generation systems. Panoptic segmentation combines the best of two worlds: Each pixel in an image is assigned a semantic class label (due to semantic segmentation) and a unique instance identifier (due to instance segmentation). It annotates both objects and the environment layout, for learning 2D scene graphs.

The central idea of our Graph2Nav is to leverage and advance these 2D panoptic scene graph works \cite{PanopticSG, PanopticSGL} into the 3D world, for generating layered 3D object-relation graphs across both indoor and outdoor environments.

\subsection{3D Scene Graphs to Navigation}

Recent works have successfully used 3D scene graphs for navigation \cite {SayPlan,SayNav,GraphPolicy,GraphMapper}. The 3D scene graphs are built either prior to the mission \cite {SayPlan} or during exploration \cite {SayNav,GraphPolicy,GraphMapper}. However, all these works are conducted at indoor environments under simulation. 

Our goal is to advance and generalize 3D scene graphs for autonomous navigation in the real world. Therefore, we integrate Graph2Nav with SayNav \cite {SayNav} on a real UGV for experiments. We also evaluate the impact from object-relation modeling, which does not exist in the 3D scene graphs from prior works\cite {SayPlan,SayNav,GraphPolicy,GraphMapper}, to navigation tasks. 


\section{Graph2Nav}
\label{sec::system}
Graph2Nav (Figure \ref{fig::system-arch}) includes three major components: (1) A SLAM system that maps and labels 3D objects in the surrounding environment using sensors (including a camera) on a mobile platform, (2) A 2D papnoptic scene graph generation module that builds a 2D object-relation graph from each keyframe during navigation, and (3) A 3D scene graph generation module that continuously merges 2D object-relation graphs into a global 3D scene graph. The details of each component are as follows.

\subsection{3D Semantic Object Extraction}
\label{sec::system-object-extraction}


We assume that a sensor system, which is composed of an RGBD camera or a LiDAR-camera suite, is equipped on a mobile platform (a UGV in our case) to provide real-time 2D RGB image and 3D depth data of the perceived environment during navigation. The depth data can be in the format of either depth image or 3D point cloud as they can be effortlessly converted to each other. For consistency, we use 3D point cloud to represent the depth data throughout the paper. We also assume the sensor system is calibrated beforehand - both system intrinsics and extrinsics parameters are available. Therefore, Graph2Nav can establish pixel-to-point relations between 2D RGB image and 3D point cloud, and vice versa for point-to-pixel relations.

Graph2Nav uses a pose graph-based SLAM system to register the received sensor data for providing real-time pose estimations of the host platform during navigation. Note Graph2Nav is designed to support various types of pose graph-based SLAM systems, whether it is vision-based, LiDAR-based, or a tightly-coupled LiDAR-vision system. For illustrative purpose, we use the factor graph formulation from LIO-SAM \cite{LIOSAM} throughout this paper. 

The factor graph is composed of three main components: odometry factor, loop closure factor, and keyframe poses $\mathbf{x}$. The RGB image and the 3D point cloud that are associated with keyframe $\mathbf{x}_i$ are denoted as $\mathbb{I}_i$ and $\mathbb{P}_i$, respectively. Note that instead of processing all received images and point clouds for the downstream modules, Graph2Nav only processes those that are associated with the keyframes of pose graph, which is not only computationally efficient but also corrects their registered poses upon the occurrence of loop closure in the SLAM system. 

The latest image $\mathbb{I}_i$ and point cloud $\mathbb{P}_i$ that are associated with the keyframe $\mathbf{x}_i$ of the SLAM system are then processed by our 2D panoptic scene graph generation module (Section \ref{sec::2DSGG}). From this 2D panoptic scene graph generation module, Graph2Nav obtains a pixel-level panoptic segmentation image $\mathbf{I}_i$ on $\mathbb{I}_i$. For each 3D point in $\mathbb{P}_i$, Graph2Nav finds its corresponding pixel in $\mathbb{I}_i$ utilizing the point-to-pixel relations and assigns it the same object label from $\mathbf{I}_i$ to obtain the semantically-labeled point cloud $\mathbf{P}_i$.

Graph2Nav then voxelizes the 3D space and assign each voxel a semantic label using each point in $\mathbf{P}_i$. Note that due to semantic label error introduced by the segmentation network, there might be voxels with different semantic labels located in it. We perform Bayesian updates for each voxel location to achieve label consistency. The semantically labeled voxel is the fundamental component of an object node in the 3D scene graph. Also, depending on the source of depth data, voxelization may greatly reduce the computation burden in this process. At last, we perform euclidean distance clustering using voxels that share the same label from multiple keyframes to extract object nodes $\mathbf{O}$. For example, an object node that describes a desk in 3D space can be composed of voxels that are labeled as desk observed from multiple semantic point clouds $\mathbf{P}$. 

In the event of loop closure, Graph2Nav reconstructs the object nodes $\mathbf{O}$ that are effected by SLAM system's pose corrections. It first finds the effected 3D semantic point clouds $\mathbf{P}$ that are associated with the corrected keyframes $\mathbf{x}$. Then, it updates the labels of voxels that are previously updated by $\mathbf{P}$. At last, it performs euclidean distance clustering on the effected voxels again to reconstruct object nodes $\mathbf{O}$. 

\subsection{2D Panoptic Scene Graph Generation}
\label{sec::2DSGG}
Graph2Nav generates a 2D object-relation graph from a RGB image $\mathbb{I}_i$ associated with each keyframe. The generation task aims to model the following distribution.

\[Pr(G|\mathbb{I}))=Pr(M,Q,R|\mathbb{I})\]

where $\mathbb{I}$ is the input image of size \textit{H} by \textit{W}. \textit{G} is the desired 2D object-relation graph which comprises a set of object masks \(M=\{m_1,m_2,...,m_n\}\) where \(m_i \in \{0,1\}^{H\times W}\), labels \(Q=\{q_1,q_2,...,q_n\}\), and object relations \(R=\{r_1,r_2,...,r_l\}\). Note each of \textit{n} object mask \(m_i\) is associated with a label \(q_i\) consisting both its class and instance ID.   

Graph2Nav adopts the one-stage PSGFormer network from \cite{PanopticSG} for this 2D object-relation graph generation task. The PSGFormer network combines the panoptic segmentation process and the 2D object-relation generation into a single end-to-end neural network architecture. It separately models the objects and relations in the form of queries from two Transformer decoders, followed by a prompting-like relation-object matching mechanism. A final prediction block simultaneously generates the object masks \textit{M} with labels \textit{Q} and their relations \textit{R}. 

To ensure effective real-time operation from the 2D panoptic scene graph (PSG) generation process, we modify the PSGFormer network backbone in \cite{PanopticSG} with a more efficient and advanced Mask2Former Vision Transformer Tiny (ViT-Tiny) backbone. We train this network using the PSG dataset proposed in \cite{PanopticSG}. Therefore, the 2D panoptic scene graphs generated from our PSGFormer model share the same 133 object classes and 56 object relationships (as in PSG dataset) across both indoor and outdoor scenes. The panoptic segmentation image $\mathbf{I}_i$ used in the real-time 3D semantic object extraction process (Section \ref{sec::system-object-extraction}) is formed by combining \textit{M} and \textit{Q} outputted from our PSGFormer.  

\subsection{3D Scene Graph Generation}
\label{sec::3DSGG}

Graph2Nav leverages the 3D semantic object management mechanism inside our SLAM system (Section \ref{sec::system-object-extraction}) to continuously merge each newly generated 2D object-relation graph \(G_i\) into a consistent global 3D scene graph $\mathbb{G}$. First, each \(G_i\) is translated into the 3D world by using the pixel-to-point relations with point cloud $\mathbb{P}_i$ that are associated with the original keyframe $\mathbf{x}_i$. We also use 3D object set $\mathbf{O}$, which is formed using information from past keyframes, to verify the object labels in \(G_i\). Inconsistent labels and masks will be corrected and updated. This improves the temporal consistency in our 3D scene graph generation process.

The 3D-translated \(G_i\) is then merged into a global 3D scene graph $\mathbb{G}$. For each new \(G_i\) from a keyframe, we compare the nodes in the 3D-translated \(G_i\) with the nodes in $\mathbb{G}$. New nodes are added to $\mathbb{G}$ with the corresponding edges from \(G_i\). We will also update existing nodes in $\mathbb{G}$ using correspondent information from \(G_i\). 

The 3D global scene graph $\mathbb{G}$ is also accumulated and arranged into a layered structure based on our definition in Figure \ref{fig::system-arch}. Same as \cite{3DSG19, Kimera, Hydra, 3DSG, SGraph, OpenGraph}, we manually define the layers in the 3D scene graph. Our definition aims to find a general and consistent hierarchy for both indoor and outdoor scenes. We define the region level (indoor rooms, outdoor areas) using specific landmarks (such as doors for separating rooms and roads for dividing areas). In the future, we will also explore the use of LLMs \cite{3DSGOntologies} to define consistent spatial ontologies. 


\section{Integration with SayNav}
\label{sec::SayNav}

We also integrate Graph2Nav with SayNav \cite{SayNav} on a real UGV for conducting navigation tasks in real environments. SayNav is a LLM-assisted hierarchical planner to search tasks. It includes three modules: (1) Incremental Scene Graph Generation, (2) High-Level LLM-based Dynamic Planner, and (3) Low-Level Planner. The High-Level LLM-based Dynamic Planner continuously converts relevant information from the 3D scene graph into text prompts to a pre-trained LLM, for dynamically generating short-term high-level plans to search target objects. Each LLM-planned step is executed by the Low-Level Planner to generate a series of control commands for execution during navigation. 

The 3D scene graph in SayNav includes four levels for indoor scenes (from bottom to top), which is compatible to our definition: small objects, large objects, rooms (regions), and house (test site). Each object node in SayNav is associated with its 3D coordinate and room node is associated with its bounds. Every door is treated as an edge to separate two rooms, which also has an associated 3D coordinate. However, all other edges in the graph only reveal the topological relationships among semantic concepts across different levels (such as a chair ``inside" a living room).  

Note SayNav is implemented under simulation. It uses ground-truth semantic labels from simulated RGBD data to build the 3D scene graph. To accomplish SayNav in the actual physical world, we use Graph2Nav to replace the original scene graph generation module in SayNav. This way also generalizes SayNav's graph generation capabilities to outdoor scenes. The generated graph also includes edges (such as ``beside" and ``on top of") to descibe relationships among objects within the same level. We also implement a classical low-level planner \cite{planner} with our UGV to support SayNav's high-level LLM-based dynamic planner.    


\section{Experiments}

In this section, we describe our experiments to evaluate Graph2Nav from two perspectives. First, we validate the accuracy of our generated 3D scene graphs for both indoor and outdoor scenes in the real world. Second, we demonstrate the impact of Graph2Nav to ground LLMs for navigation in real unknown environments.

\subsection{Experimental Setup}
\label{sec::sensor}

Our sensor system for Graph2Nav includes two sensors: a Livox Mid-360 3D LiDAR and a Realsense D455 camera. Both sensors are low-cost and have integrated IMU which provides us with the capability of conducting challenging experiments in indoor and outdoor environments under aggressive motions. The 3D LiDAR has a detection range of 40 meters and a field-of-view (FOV) of 360-by-59 degrees. The Realsense camera has one RGB and two infrared imaging sensors, which have a FOV of 90-by-65 degrees. Our sensor system is insensitive to the platform, which enables us to use the same setup with a variety of platforms, including handheld devices, ground robots, legged robots, and drones.

We use LIO-SAM \cite{LIOSAM} as the pose graph-based 3D SLAM system in Graph2Nav. We also modified the PSGFormer network \cite{PanopticSG} to ensure real-time operation to object-relation graph generation. Our system can recognize 133
object classes and 56 object relationships across both indoor and outdoor scenes. These relationships (edges) describe either spatial relations (such as ``in front of" and ``beside"), states of one object related by another (such as a person ``sits on" a chair), or semantic relations that are expressed by prepositions (e.g., ``with"). The details are in Section \ref{sec::system}.  

\subsection{Accuracy of 3D Object-Relation Graph Generation}

To evaluate the quality and generality of our 3D object-relation graph, we collected data using our sensor system in four different environments (two for indoors, and two for outdoors) for experiments. The goal is to validate our generated 3D scene graphs to a wide variety of objects within different indoor and outdoor scenes. We collected the first indoor dataset within a large cafeteria and nearby rest areas. The cafeteria includes dining tables and chairs inside the scenes, while there are sofas and equipment (such as a table tennis table) in the rest area. The second environment is our laboratory space, including typical office objects such as desks, monitors, books, and computers. For the first outdoor environment, we explore a large courtyard, which includes natural objects such as trees and bushes. The second outdoor dataset is a large parking lot and nearby roads. It includes typical objects inside urban scenes, such as cars and poles. Datasets from these environments are ranged roughly from a half kilometer to a kilometer in total trajectory length, except the second indoor environment (around 250 meters).  


We measure 3D coordinates of a set of representative 3D objects using state-of-the-art survey techniques inside these three environments. There are total 67 measured objects: 32 objects in cafeteria, 38 objects in lab, 22 objects in courtyard, and 13 objects in parking lot. We use these measured objects as ground truth, for validating the 3D scene graphs generated from Graph2Nav. Note the accuracy of our mapped 3D point cloud and the estimated platform pose relies on the underlying SLAM system (LIO-SAM \cite{LIOSAM} in our case), which is not the focus of our work.   


We apply Graph2Nav on the datasets collected from these three environments. We then compare the estimated 3D object coordinates (centroids) against the ground truth. Table \ref{tbl:sgaccuracy1} shows the average 3D error in estimating the centroids of 3D objects among three different environments. The second row reports the 3D object localization error based on the 3D point clouds correspondent to a single 2D segmented image produced by \cite{PanopticSG}. The third row describes errors from our proposed graph generation framework. It shows Graph2Nav reduces object localization error across a wide variety of objects from three environments, by leveraging 3D SLAM techniques to improve temporal consistency across 2D object segmentations from sequential video frames. 

\begin{table}[t]
    \centering
    \caption{Accuracy of 3D Objects (meters, the lower the better)}
    \begin{tabular}{c|c|c|c|c}
        \hline
        \textbf{\begin{tabular}[c]{@{}c@{}} \small  \end{tabular}} & \small \textbf{Indoor1}
        & \small \textbf{Indoor2}
        & \multicolumn{1}{c|}{\small \textbf{Outdoor1}} & \multicolumn{1}{c}{\small \textbf{Outdoor2}}  \\
        \hline
        \hline
        \textbf{\begin{tabular}[c]{@{}c@{}} \small Single Image \\ \small (baseline)\end{tabular}} & \small 0.5898 & \small 0.2530 & \small 0.5755 & \small 1.0403  \\
        \hline
        \textbf{\begin{tabular}[c]{@{}c@{}} \small Graph2Nav \\ \small (ours)\end{tabular}} & \small 0.3867 & \small 0.2236 & \small 0.3619 & \small 0.8407  \\
        \hline
        \textbf{\begin{tabular}[c]{@{}c@{}} \small Our \\ \small Improvement \end{tabular}} & \small 34.44\% & \small 11.62\% & \small 37.12\% & \small 19.19\%  \\
        \hline
    \end{tabular}
    \vspace{-0.7em}
    \label{tbl:sgaccuracy1}
\end{table}

We also manually label the object relations, based on the definition in \cite{PanopticSG}, among the representative objects inside the collected datasets. Note only a small subset from all possible object pairs can have meaningful semantic relationships (edges). For examples, trees and bushes spread out in the courtyard. Therefore, there are very few valid relations (edges) among these objects, which are well separated in space. For effective evaluation, we combine 4 environments into two categories: indoors and outdoors. We annotate 30 indoor and 17 outdoor valid object-relations as ground truth.


Table \ref{tbl:sgaccuracy2} shows the successful rate in labeling correct relations among objects in both indoor and outdoor environments. The second row reports the successful rate of object relations detected from one 2D panoptic graph produced by \cite{PanopticSG}. The third row represents the successful rate from Graph2Nav, which adapts and advances 2D panoptic graph into the 3D world using the techniques described in Section \ref{sec::3DSGG}. Clearly, Graph2Nav greatly improves the capability in detecting correct relationships among the objects, compared to 2D-based method. Note the actual objects in the 3D world can be perceived from a wide range of viewpoints and distances. Combining observations from different 2D images shall improve the robustness and accuracy in depicting the actual relationships among 3D objects in the real world.   

\begin{table}[t]
    \centering
    \caption{Accuracy of Relations (percentage, the higher the better)}
    \begin{tabular}{c|c|c}
        \hline
        \textbf{\begin{tabular}[c]{@{}c@{}} \small  \end{tabular}} & \small \textbf{Indoor} & \multicolumn{1}{c}{\small \textbf{Outdoor}} \\
        \hline
        \hline
        \textbf{\begin{tabular}[c]{@{}c@{}} \small 2D Graph \\ \small (baseline)\end{tabular}} & \textbf{\begin{tabular}[c]{@{}c@{}} \small 56.67\% \\ \small (17/30)\end{tabular}} &  \textbf{\begin{tabular}[c]{@{}c@{}} \small 64.71\% \\ \small (11/17)\end{tabular}}  \\
        \hline
        \textbf{\begin{tabular}[c]{@{}c@{}} \small Graph2Nav \\ \small (ours)\end{tabular}} &  \textbf{\begin{tabular}[c]{@{}c@{}} \small 83.33\% \\ \small (25/30)\end{tabular}}  &  \textbf{\begin{tabular}[c]{@{}c@{}} \small 94.12\% \\ \small (16/17)\end{tabular}}   \\
        \hline
        \textbf{\begin{tabular}[c]{@{}c@{}} \small Our \\ \small Improvement \end{tabular}} &  \textbf{\begin{tabular}[c]{@{}c@{}} \small 26.66\% \end{tabular}}  &  \textbf{\begin{tabular}[c]{@{}c@{}} \small 29.41\% \end{tabular}}   \\
        \hline
    \end{tabular}
    \vspace{-2em}
    \label{tbl:sgaccuracy2}
\end{table}

\begin{figure}[t]
	\centering
	\includegraphics[width=.48\textwidth]{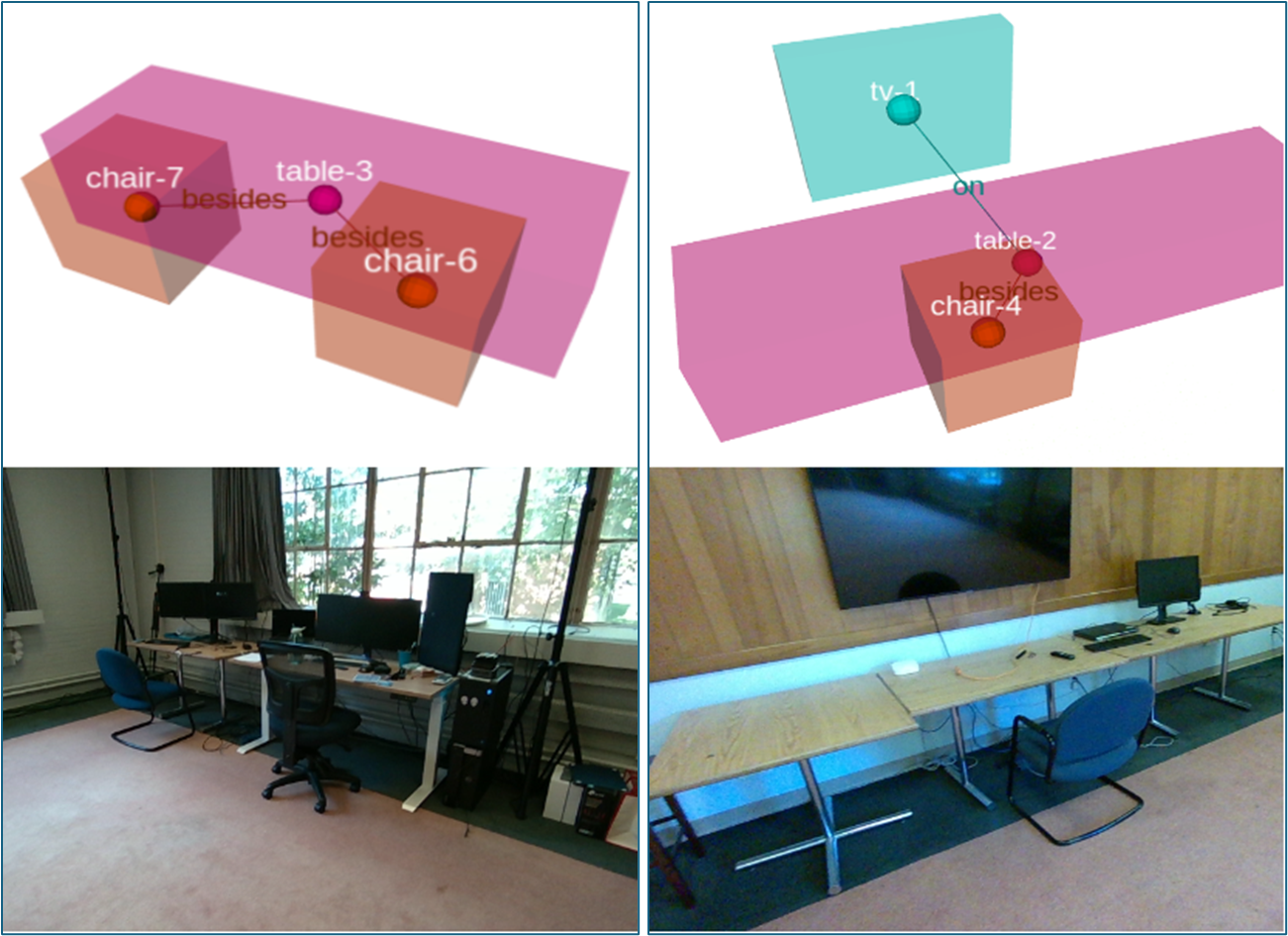}
	\caption{Two examples of object-relations ("beside" and "on" top of) from portions of our generated 3D scene graphs (top) with their correspondent 2D images (bottom). Note we only show large objects with their relations for better visualization.}
	\label{fig::relation}
	\vspace{-5mm}
\end{figure}

Figure \ref{fig::relation} shows two examples of object relations generated from Graph2Nav. Each object instance is labeled (white text) with its object class and instance ID. The 3D bounding box for the object instance is also displayed in our graph. The relations among objects are shown as edges with labels (red or green texts). The visualization in the left example shows two chairs are placed "beside" (nearby) the same table. The right example shows a chair is placed "beside" a table while a TV is on "top" of the same table. 

\subsection{Impact to Robot Navigation}

To evaluate the impact from Graph2Nav to robot navigation, we integrate Graph2Nav with SayNav on a UGV, as described in Section \ref{sec::SayNav}.
Our UGV, Rover Zero 3, is a ground robot and equipped with all-wheel drive. The ruggedized wheels of the robot enables it to operate on various types of terrains and is capable of supporting a maximum load of 75Kg. With a battery, the robot can operate continuously for up to 2 hours. We installed our sensor system (Section \ref{sec::sensor}) on top of this robot. We use Nvidia AGX Orin as our onboard computer for Graph2Nav processing and SayNav inference. 

We integrate our entire autonomous navigation stack based on Robot Operating System 2 (ROS2). With ROS2's Data Distribution Serivce (DDS), we can communicate different processing modules either locally or remotely. We also use OpenAI's Whisper for speech-to-text translation. The user can therefore assign a navigation task (such as finding a backpack) to the robot via voice commands. The translated text are then parsed into SayNav and processed by GPT 3.5 (a LLM used by SayNav), for executing the task.

During execution of the navigation task, the sensor data received from our sensor system is continuously fed into Graph2Nav, for generating 3D scene graphs in real-time during exploration. SayNav then feeds these scene graphs into remote GPT 3.5, for generating high-level step-by-step plans (such as moving to object A, looking around). We use Nav2 \cite{planner}, which provides path planning and low-level control, to execute each planned step from the LLM in SayNav. 

We evaluate the impact from Graph2Nav via SayNav to object search tasks in a real unknown environment. The approach of SayNav to object search tasks is to use LLMs to generate a search plan based on the perceived environment (3D scene graphs as inputs to LLMs). The generated plan from LLMs provides an efficient search strategy within the current perceived area based on human knowledge, prioritizing locations to visit inside the room based on the likelihoods of target objects being discovered. For example, LLM may provide a plan to first check the desk and then the bed to find the laptop in the bedroom. 

We use our laboratory space as the unknown environment for evaluating these search tasks. We conducted six search scenarios for evaluation. For each scenario, we have different arrangements of 13 large objects (tables, chairs, sofas etc) and many small objects inside the environment. We command the robot to search a single small object (target object), either a water bottle or a backpack, for each scenario. Note the environment is unknown to the robot prior to the scenario. Therefore, once the robot starts the task, it will first look around to build the initial scene graph of the perceived environment using Graph2Nav. The scene graph includes mostly large objects due to the perceived distances (many small objects in our lab cannot be detected from far-away distances). The robot then will set up and execute a search plan for finding the target object based on LLM's knowledge of the perceived environment. The scene graph will be expanded and augmented during navigation. The plan can also be dynamically changed, updated, or re-planned during execution, if any failure happens or any new information is received. More details on how SayNav works in object search tasks can be found in \cite{SayNav}.

\begin{table}[t]
    \centering
    \caption{Search Time (seconds, the lower the better)}
    \begin{tabular}{c|c|c|c}
        \hline
        \textbf{\begin{tabular}[c]{@{}c@{}} \small Target \\
        \small Object \end{tabular}} & \begin{tabular} [c]{@{}c@{}} \small \textbf{No Relations} \\ \small \textbf{\cite{SayNav}} 
        \end{tabular} & \begin{tabular} [c]{@{}c@{}} \small \textbf{With Relations} \\ \small \textbf{(ours)} 
        \end{tabular} & \begin{tabular} [c]{@{}c@{}} \small \textbf{Our} \\ \small \textbf{Improvement} 
        \end{tabular} \\
        \hline
        \hline
        \textbf{\begin{tabular}[c]{@{}c@{}} \small Backpack\end{tabular}} & \small 165.0 & \small 124.8 & \small 24.36\% \\
        \hline
        \textbf{\begin{tabular}[c]{@{}c@{}} \small Backpack\end{tabular}} & \small 184.7 & \small 147.0 & \small 20.41\% \\
        \hline
         \textbf{\begin{tabular}[c]{@{}c@{}} \small Backpack\end{tabular}} & \small 222.6 & \small 163.8 & \small 
         26.42\% \\
        \hline
        \textbf{\begin{tabular}[c]{@{}c@{}} \small Bottle\end{tabular}} & \small 250.3 & \small 163.1 & \small
        34.84\% \\
        \hline
        \textbf{\begin{tabular}[c]{@{}c@{}} \small Bottle\end{tabular}} & \small 152.6 & \small 111.4 & \small
        27.00\% \\
        \hline
         \textbf{\begin{tabular}[c]{@{}c@{}} \small Bottle\end{tabular}} & \small 166.4 & \small 136.3 & \small
         18.09\% \\
        \hline
    \end{tabular}
    \vspace{-1em}
    \label{tbl:navigationresult}
\end{table}

Table \ref{tbl:navigationresult} shows the search time for each of six scenarios. Note we count the search time until the robot detects (finds) the target object in its perceived image. For each scenario, we also evaluate the impact of the object-relations, which do not exist in \cite{SayNav}, from the 3D scene graphs generated from Graph2Nav. It means the robot does two trials (one uses the graph without relations, and the other uses the entire graph with object relations from Graph2Nav) for the same scenario. From the results, we found that the LLM is able to utilize the object-relations to design more efficient plans to search objects. Therefore, the search time for the same scenario is reduced by using the object-relations from Graph2Nav.

\begin{figure}[t]
	\centering
	\includegraphics[width=.48\textwidth]{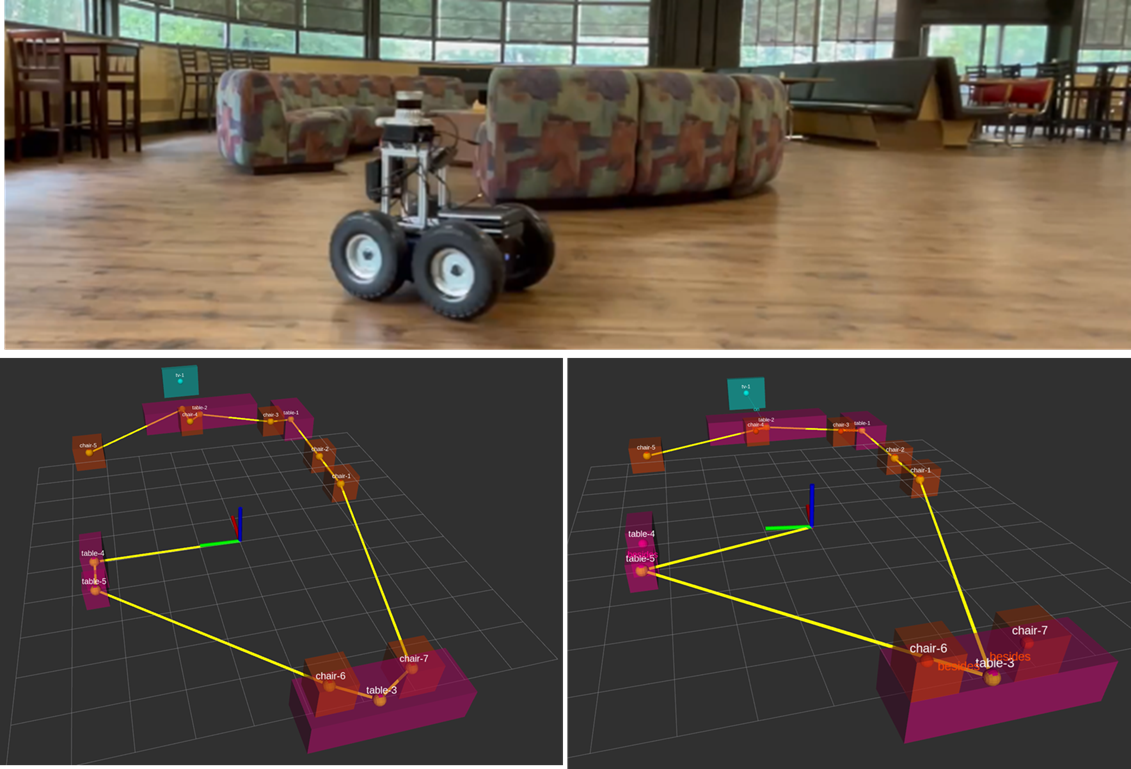}
	\caption{An example of the impact from object-relations to the search plan (yellow trajectory in bottom-left picture and bottom-right picture) executed by our robot (top): No object-relations (bottom-left), and with object-relations (bottom-right).}
	\label{fig::path}
	\vspace{-5mm}
\end{figure}

Figure \ref{fig::path} shows an example on how object-relations impacts LLMs to generate more efficient plans to search the target object using our robot. Without object-relations, the LLM generates the initial search plan based on (1) the distances to different large objects inside the environment, and (2) the likelihoods of target objects being discovered at those locations (large objects). With object-relations, the LLM also considers the third dimension - the relationships among these large objects - to design the initial search plan. As shown in Figure \ref{fig::path}, two chairs and one table are close to each other ("beside") in the bottom of the environment. The LLM shortens the path by visiting only the table, based on these relations. The robot can look around to cover the two nearby chairs, when it visits the table. Therefore, the search time is reduced.    
 
\section{Conclusions and Discussion}

Our goal is to advance and generalize 3D scene graphs for autonomous navigation in the real world. To fulfull this goal, we propose Graph2Nav, a novel real-time 3D object-relation graph generation framework that addresses current limitations to robot navigation. Our framework fully generates and exploits both 3D objects and a rich set of semantic relationships among objects in a 3D scene graph, which is applicable to both indoor and outdoor scenes. To avoid previous training data constraint in learning 3D scene graphs, Graph2Nav leverages and advances existing 2D object-relation graph works into the 3D world. We validate the accuracy improvement from 3D mapping in scene graph generation. We also evaluate the impact from Graph2Nav to robot navigation via integration of SayNav on a UGV to object search tasks in real unknown environments.

Future work is to further exploit the object-relations in 3D scene graphs to more complicated robotic applications. For example, grounding these object-relations in LLMs shall enable understanding of fine-grained relationships among objects and generate efficient plans to wide-area manipulation tasks (such as rearranging inventory inside a factory). 

\newpage


%


\vspace{-1mm}

\begin{thebibliography}{99}


\bibitem{3DSG19}
I. Armeni, Z. He, J. Gwak, A. Zamir, M. Fischer, J. Malik, and S. Savarese, ``3d scene graph: A structure for unified semantics, 3d space, and camera", \textit{IEEE International Conference on Computer Vision and Pattern Recognition}, 2019.


\bibitem{Kimera}
A. Rosinol, A. Violette, M. Abate, N. Hughes, Y. Chang, J. Shi, A. Gupta, and L. Carlone, ``Kimera: From slam to spatial perception with 3d dynamic scene graphs", \textit{The International Journal of Robotics Research}, vol. 40, no. 12-14, pp. 1510-1546, 2021.
 
\bibitem{Hydra}
N. Hughes, Y. Chang, and L. Carlone, ``Hydra: A real-time spatial perception engine for 3d scene graph construction and optimization", \textit{Robotics: Science and Systems}, 2022.

\bibitem{3DSG}
A. Rosinol, A. Gupta, M. Abate, J. Shi, and L. Carlone, ``3D Dynamic Scene Graphs: Actionable Spatial Perception with Places, Objects, and Humans", \textit{Robotics: Science and Systems}, 2020. 

\bibitem{SGraph}
H. Bavle, J. Sanchez-Lopez, M. Shaheer, J. Civera, and H. Voos, ``S-Graphs+: Real-Time Localization and Mapping Leveraging Hierarchical Representations", \textit{IEEE Robotics and Automation Letters}, vol. 8, pp. 4927-4934, 2023. 

\bibitem{OpenGraph}
Y. Deng, J. Wang, J. Zhao, X. Tian, G. Chen, Y. Yang, and Y. Yue, "OpenGraph: Open-Vocabulary Hierarchical 3D Graph Representation in Large-Scale Outdoor Environments", \textit{IEEE Robotics and Automation Letters}, vol. 9, pp. 8402-8409, 2024. 

\bibitem{3DSGOntologies}
J. Strader, N. Hughes, W. Chen, A. Speranzon, and L. Carlone, "Indoor and Outdoor 3D Scene Graph Generation via Language-Enabled Spatial Ontologies", \textit{IEEE Robotics and Automation Letters}, vol. 9, pp. 4886-4893, 2024. 

\bibitem{HSG2Robotics}
N. Hughes, Y. Chang, S. Hu, R. Talak, R. Abdulhai, J. Strader, and L. Carlone, "Foundations of Spatial Perception for Robotics: Hierarchical Representations and Real-Time Systems", \textit{The International Journal of Robotics Research}, 2024. 

\bibitem{Wald20}
J. Wald, H. Dhamo, N. Navab, and F. Tombari, "Learning 3D semantic scene graphs from 3D indoor reconstructions", \textit{IEEE International Conference on Computer Vision and Pattern Recognition}, 2020.
  
\bibitem{Wu21}
S. Wu, J. Wald, K. Tateno, N. Navab, F. Tombari, "SceneGraphFusion: Incremental 3D scene graph prediction from RGB-D sequences",\textit{IEEE International Conference on Computer Vision and Pattern Recognition}, 2021.

\bibitem{Wu23}
S. Wu, K. Tateno, N. Navab, F. Tombari, "Incremental 3D Scemantic Scene Graph Prediction from RGB Sequences",\textit{IEEE International Conference on Computer Vision and Pattern Recognition}, 2023.



\bibitem{taskography}
C. Agia et al., "Taskography: Evaluating robot task planning over large 3D scene graphs", ,\textit{Proceeding of the 5th Conference on Robot Learning}, pp. 46-58, 2022.

\bibitem{imagemanipulation}
H. Dhamo, A. Farshar, I. Laina, N. Navab, G. Hager, F. Tombari, and C. Rupprecht, "Semantic Image Manipulation Using Scene Graphs", \textit{IEEE International Conference on Computer Vision and Pattern Recognition}, 2020.

\bibitem{GraphTo3D}
H. Dhamo, F. Manhardt, N. Navab, and F. Tombari, "Graph-to-3d: End-to-end generation and manipulation of 3d scenes using scene graph", \textit{IEEE/CVF International Conference on Computer Vision}, pp. 16352-16361, 2021.

\bibitem{SayPlan}
K. Rana, J. Haviland, S. Garg, J. Abou-Chakra, I. Reid, and N. Suenderhauf, "Grounding Large Language Models using 3D Scene Graphs for Scalable Robot Task Planning", \textit{Conference on Robot Learning}, 2023.

\bibitem{SayNav}
A. Rajvanshi, K. Sikka, X. Lin, B. Lee, H. Chiu, and A. Velasquez, "SayNav: Grounding Large Language Models for Dynamic Planning to Navigation in New Environments", \textit{International Conference on Automated Planning and Scheduling}, 2024.

\bibitem{GraphPolicy}
Z. Ravichandran, L. Peng, N. Hughes, J. Griffith, and L. Carlone, "Hierarchical Representations and Explicit Memory: Learning Effective Navigation Policies on 3D Scene Graphs using Graph Neural Networks", \textit{International Conference on Robot Automation}, 2022.

\bibitem{GraphMapper}

Z. Seymour, N. Mithun, H. Chiu, S. Samarasekera, and R. Kumar,
"GraphMapper: Efficient Visual Navigation by Scene Graph Generation", \textit{International Conference on Pattern Recognition}, 2022.

\bibitem{SGSurvey}
G. Zhu, L. Zhang, Y. Jiang, Y. Dang, H. Hou, P. Shen, M. Feng, X. Zhao, Q. Miao, S. Shah and M. Ben, "Scene Graph Generation: A Comprehensive
Survey", \textit{Neurocomputing}, vol. 566, 2024.

\bibitem{PanopticSG}
J. Yang, Y. Ang, Z. Guo, K. Zhou, W. Zhang, and Z. Liu, "Panoptic Scene Graph Generalization", \textit{European Conference on Computer Vision}, 2022.

\bibitem{PanopticSGL}
L. Li, W. Ji, Y. Wu, M. Li, Y. Qin, L. Wei, R. Zimmermann, "Panoptic Scene Graph Generation with Semantics-Prototype Learning", \textit{Annual AAAI Conference on Artificial Intelligence}, 2024.

\bibitem{PanopticSegmentation}
A. Kirillov, K. He, R. Girshick, C. Rother, and P. Dollar, "Panoptic Segmentation", \textit{IEEE International Conference on Computer Vision and Pattern Recognition}, 2019.

\bibitem{LIOSAM}
T. Shan, B. Englot, D. Meyers, W. Wang, C. Ratti, and D. Rus, "LIO-SAM: Tightly-coupled Lidar Inertial Odometry via Smoothing and Mapping", \textit{IEEE/RSJ International Conference on Intelligent Robots and Systems}, 2020.

\bibitem{planner}
S. Macenski, F. Martín, R. White, and J. Clavero, "The Marathon 2: A Navigation System", \textit{IEEE/RSJ International Conference on Intelligent Robots and Systems}, 2020.

\end{thebibliography}
\end{document}